\documentclass{article}

\usepackage[preprint,nonatbib]{neurips_2026}

\usepackage[utf8]{inputenc} 
\usepackage[T1]{fontenc}    
\usepackage{hyperref}       
\usepackage{url}            
\usepackage{booktabs}       
\usepackage{amsfonts}       
\usepackage{nicefrac}       
\usepackage{microtype}      
\usepackage{xcolor}         

\usepackage{graphicx}
\usepackage{enumitem}
\usepackage{makecell}
\usepackage{amsmath}
\usepackage{multirow}
\usepackage{algorithm}
\usepackage{booktabs}
\usepackage[table]{xcolor}
\usepackage{listings}
\usepackage{algpseudocode}
\usepackage{subcaption}

\usepackage{amssymb}
\usepackage{hyperref}
\usepackage{float}
\usepackage{bigstrut}
\usepackage{tabularx}
\usepackage{wrapfig}

\title{NTS-CoT: Mitigating Hallucinations in LLM-based News Timeline Summarization with Chain-of-Thought Reasoning}

\author{
    \parbox{\textwidth}{\centering
    Feng Lyu\textsuperscript{1},
    Huiqin Yan\textsuperscript{1},
    Sijing Duan\textsuperscript{2},
    Hao Wu\textsuperscript{3},
    Shuang Gu\textsuperscript{4},
    Xue Qiao\textsuperscript{4},
    Weixu Zhang\textsuperscript{5},
    Haolun Wu\textsuperscript{5}
    }
    \\[1ex]
    \textsuperscript{1}Central South University,
    \textsuperscript{2}Tsinghua University,
    \textsuperscript{3}Nanjing University,
    \\
    \textsuperscript{4}Suzhou Aerospace Information Research Institute,
    \textsuperscript{5}McGill University
}

\begin{document}

\maketitle

\begin{abstract}
  The rapid updates of online news make tracking event developments challenging, highlighting the need for timeline summarization (TLS). Hallucinations, where LLM-generated content deviates from source news, still remain a critical issue in LLM-based TLS and are not well studied in existing works. To bridge this gap, we identify two primary types of hallucinations: unfaithful content during news summarization and information omission in date-event summarization. Then, we propose \texttt{NTS-CoT}, a novel framework that leverages Chain-of-Thought (CoT) reasoning to mitigate hallucinations in TLS. The framework consists of three key modules: i) \emph{Element-CoT} to capture essential news elements for faithful summarization, ii) \emph{Date Selection} to combine temporal saliency and event prominence for timestamp selection, and iii) \emph{Causal-CoT} to infer causal relationships and reduce omissions in date-event summarization. Extensive experiments, including quantitative analysis on three TLS benchmarks and human evaluation, demonstrate that \texttt{NTS-CoT} outperforms state-of-the-art baselines, effectively mitigating hallucinations and improving LLM-based TLS performance. Our source code is available at \url{https://anonymous.4open.science/r/NTS-CoT}.
\end{abstract}

\section{Introduction}

The rapid proliferation of news on the Internet, characterized by its complexity and continuous updates, has made it increasingly difficult for individuals to track and comprehend the progression of events efficiently. Timeline summarization (TLS) addresses this challenge by organizing pivotal stages of a topic's development chronologically and providing concise summaries for each timestamp. 
By offering a structured overview of historical event developments, TLS helps users quickly identify the trends and patterns surrounding their query topic.
This capability is especially valuable in critical contexts such as disaster response, policy evolution, and trend analysis.

Traditional TLS research mainly focuses on date selection \cite{10.1145/2487788.2487829,tran2015joint,gholipour-ghalandari-ifrim-2020-examining} and extractive event summarization \cite{chieu2004query,martschat-markert-2018-temporally}, but sentence-level extraction may suffer from content overlap and redundancy, logical discontinuity, or details omission, thereby reducing the readability of timeline summaries. Recently, large language models (LLMs) have demonstrated strong capabilities in TLS\cite{brown2020language,zhang2023extractive}. However, LLMs are prone to hallucinations \cite{ye2022unreliability,tam-etal-2023-evaluating,ji2023survey,hase2023methods,thakur2025assessing}, where generated content may appear fluent and coherent but fail to align with actual events or context. Existing LLM-based studies mainly define new TLS tasks and apply LLMs to them, with limited research on hallucination mitigation\cite{hu-etal-2024-moments,wang2023web}.

\begin{wrapfigure}{r}{0.5\textwidth}
\vspace{-0.6cm}
  \centering
  \includegraphics[width=\linewidth]{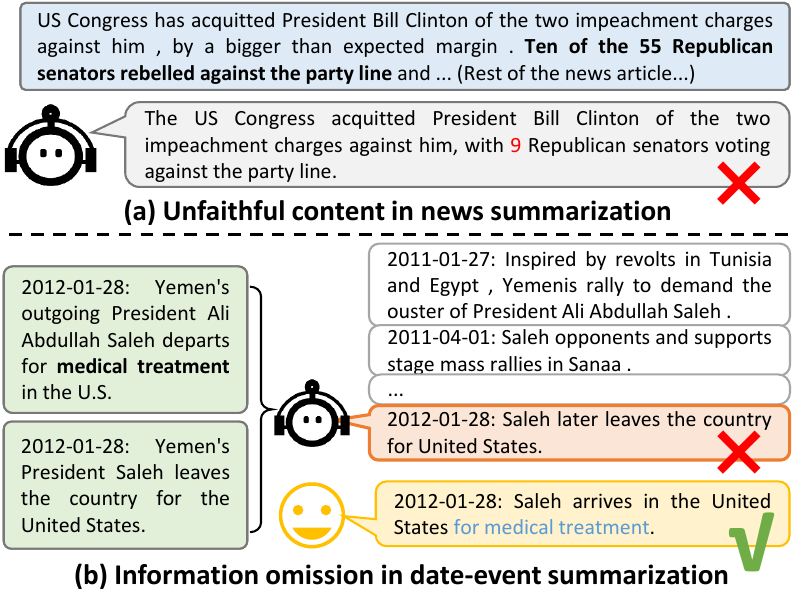}
  \caption{Examples of LLMs' hallucinations in the news timeline summarization task.}
  \label{fig:hallucination_example}
  \vspace{-0.5cm}
\end{wrapfigure}

Hallucinations in LLM-based TLS primarily occur during the generation of news summaries and date-event summaries, encompassing both single-document summarization and multi-document summarization. As illustrated in Figure~\ref{fig:hallucination_example}, a key manifestation of hallucination is the generation of \textbf{Unfaithful Content} that deviates from the source news \cite{maynez-etal-2020-faithfulness} during the \textit{news summarization stage}. Additionally, the diversity and overlap of the news content, coupled with the interdependencies between events, often result in \textbf{Information Omission} \cite{berezin2023named,tonmoy2024comprehensive,huang2025survey} during the \textit{date-event summarization stage}. These phenomena highlight the need for more robust mechanisms to ensure accuracy and completeness in LLM-generated TLS outputs.

To mitigate hallucinations in LLM-based news timeline summarization, Chain-of-Thought (CoT) is an effective approach. CoT enhances the model's ability to generate faithful and coherent summaries by breaking complex tasks into intermediate reasoning steps, thereby reducing errors and improving alignment with source content \cite{wei2022chain,yao2023tree,wang-etal-2023-element}. However, two key challenges remain: (i) How can CoT be designed to enhance the LLM's focus on the event details and milestone date selection of the single source news to prevent the generation of unfaithful content? (ii) How can the LLM be guided to identify and retain potential relationships across multiple documents, thereby minimizing information omission?

In this paper, we propose a novel framework, \emph{NTS-CoT}, which employs well-designed CoT reasoning techniques to address these challenges. \emph{NTS-CoT} operates in three stages: single-document-based news summarization, date selection, and multi-document-based date-event summarization. In the first stage, we design an Element-CoT module, which infers key elements (e.g., event, entity, location, date, and result) from topic-related news, effectively reducing unfaithful content hallucinations by ensuring critical details are accurately captured. During date selection, we construct an associated date graph to evaluate date importance and perform event clustering to identify milestone events. Finally, in the date-event summarization stage, a Causal-CoT module is designed to reason about causal relationships between events across multiple documents, further mitigating information omission hallucinations and producing timeline summaries that are more consistent with the original content. 

We conduct quantitative analysis experiments on three TLS benchmarks and perform human evaluation to demonstrate the effectiveness of our method. Results show that \emph{NTS-CoT}  improves the AR-1 by 23.4\%, AR-2 by 33.4\% and Date-F1 by 10.0\% compared to the SOTA baseline, demonstrating \emph{NTS-CoT} excels in both textual fidelity and temporal accuracy, effectively reducing factual inconsistencies and omissions in generated timelines. Human evaluation further confirms the superiority of \emph{NTS-CoT}, with evaluators preferring its summaries in 67.74\% of cases for faithfulness and 54.38\% for completeness over LLM-TLS.
In summary, our contributions are as follows:
\begin{itemize}[leftmargin=10pt]
    \item We propose a new framework \texttt{NTS-CoT}, which mitigates two types of hallucinations in LLM-based TLS: unfaithful content hallucinations in news summarization and information omission hallucinations in date-event summarization.
    \item We introduce two key CoT modules: Element-CoT and Causal-CoT. Element-CoT addresses unfaithful content hallucinations by grounding summaries on verifiable news elements, while Causal-CoT reduces omission hallucinations by leveraging causal reasoning to integrate information across multiple documents. We also implement a date selection module balancing temporal saliency and event prominence.
    \item Quantitative analysis and human evaluation demonstrate the effectiveness of \texttt{NTS-CoT} on the TLS task, outperforming baselines on three real-world benchmark datasets. All our code is open source.
\end{itemize}

\section{Preliminaries}
\subsection{Timeline Summarization}
Given a set of news articles $A = \{a_1, a_2, \dots, a_n\}$ over a time range $\mathcal{T}=\{t_1,t_2,\dots,t_n\}$ and a set of topic query keyphrases $Q$, where each article includes a publication time. Our task is to generate a timeline summarization $S$ for each topic keyphrase that consists of $l$ timestamp $t_i \in \mathcal{T}$, each associated with an event summary $s_{t_i}$. For example, we extract a summary $s_{t_i}$ containing a timestamp $t_i$ in a fixed format: ``1999-01-07: The impact trial of President Bill Clinton begins, with the Senate set to vote on his removal from office.'' Typically, the timeline summarization is organized in chronological order and is represented as $S = \{{t_1},{s_{{t_1}}};{t_2},{s_{{t_2}}};{\dots};{t_l},{s_{{t_l}}}\}$.

\subsection{Hallucination in TLS}
Before determining hallucination types in news TLS, we conduct a preliminary evaluation. We randomly sample 1500 news and 300 date-event news from the three benchmark datasets \cite{tran2013leveraging, tran2015timeline, gholipour-ghalandari-ifrim-2020-examining}, generate summaries and ask volunteers to manually annotate hallucinations. Results show that 28.6\% of single-news summaries contain unfaithful content hallucinations, while 23.7\% of date-event summaries exhibit information omission, and these rates are deemed relatively high.

Therefore, our investigation confirms that hallucinations in LLM-based TLS primarily manifest in two stages: news summarization and date-event summarization. Hallucinations in the news summarization arise when the LLM generates an summary $\hat{s}_{a_i}$ for a given article $a_i$, containing content inconsistent with or absent from the original text (see Figure~\ref{fig:hallucination_example} (a) for an example). We define the hallucinations in the news summarization stage as the output summary $\hat{s}_{a_i}$ is not faithful to the given article $a_i$. In the date-event summarization stage, hallucinations manifest in the aggregation of multiple summaries $\{\hat{s}_{a_1},\dots \hat{s}_{a_n}\}$ into $s_{t_i}$ (see Figure~\ref{fig:hallucination_example} (b) for an example). We define the hallucinations at this stage as the summary $s_{t_i}$ output by LLM that omits important information or has redundant information.


\section{Methodology}

\begin{figure*}[h]
\centering
  \includegraphics[width=1\textwidth]{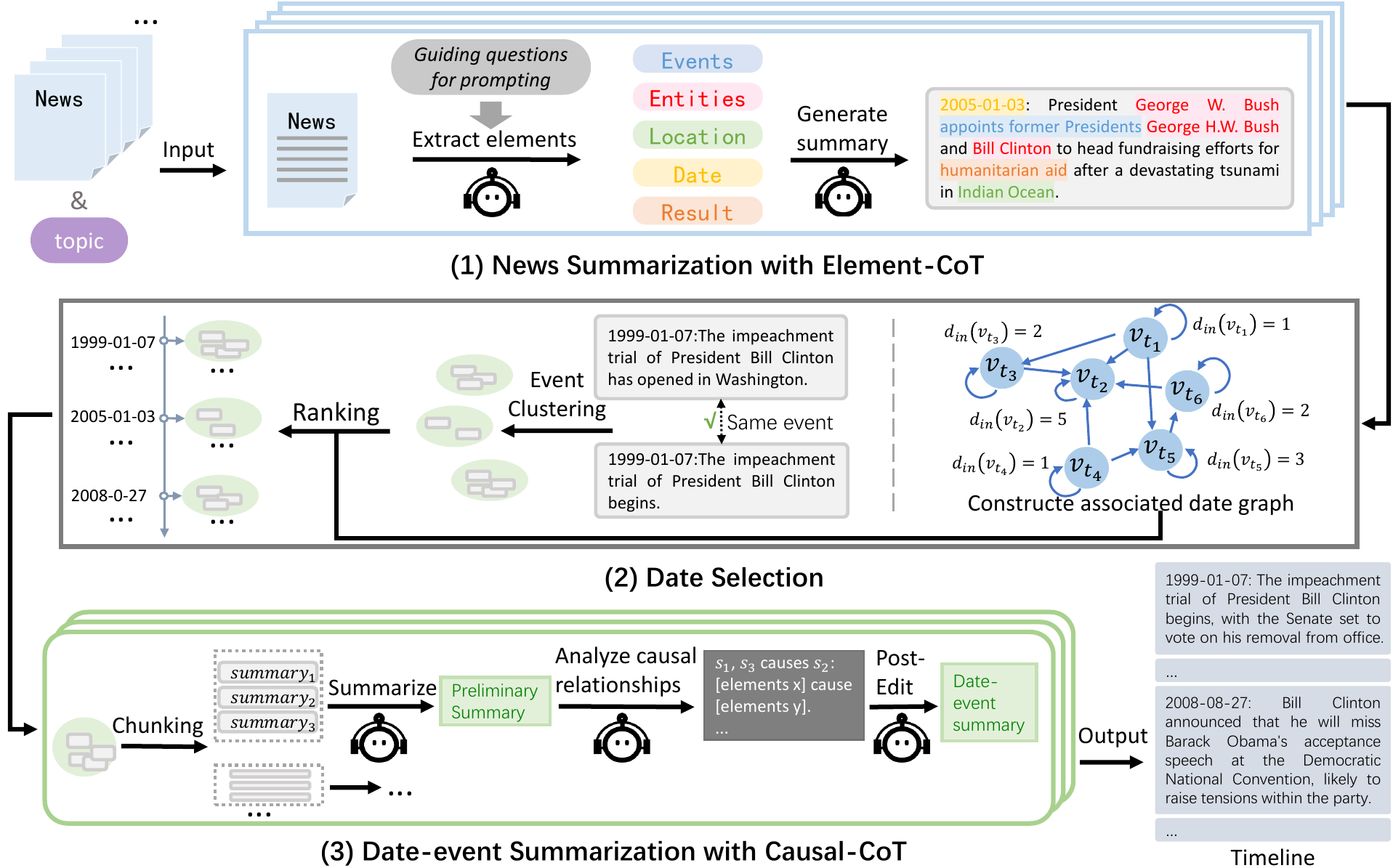}
  \caption{An overview of the \texttt{NTS-CoT} pipeline for TLS and an example. Given news articles and a topic query, the pipeline (1) summarizes each article with Element-CoT, (2) selects dates for the timeline, and (3) summarizes event clusters for those dates with Causal-CoT. Gray squares and green circles represent the news summaries and event clusters of the same events, respectively, with each cluster containing multiple news summaries describing the same event.}
  \label{fig:overview}
\end{figure*}

In this section, we propose \texttt{NTS-CoT}, a new framework that leverages CoT reasoning ability to mitigate hallucinations in LLMs for TLS tasks and achieve accurate timeline summarization for specific topics. As shown in Figure~\ref{fig:overview}, \texttt{NTS-CoT} consists of three key modules: news summarization with \emph{Element-CoT} to mitigate faithfulness hallucinations, date selection, and date-event summarization with \emph{Causal-CoT} to reduce information redundancy and omission.

\subsection{News Summarization}
\label{sub:ns}
In this section, we use LLMs to generate summaries for each news article. Previous approaches often relied on extracting representative sentences directly to serve as its summary, which guarantee surface-level faithfulness but are often incomplete and redundant, capturing only parts of an event. By contrast, generative summarization offers the ability to integrate scattered information across the article into a concise, semantically coherent description.

However, LLMs may generate unfaithful content due to confusion or conflicts between internal knowledge and the context in news summarization. To tackle this issue, we design an \emph{Element-CoT} prompt to first instruct LLMs to focus on analyzing key news elements within an article, extract structured event elements as verifiable factual units, and then condition the generative step on both these elements and the original text when summarizing each news article. 

\subsubsection{News Element Extraction}
In professional news writing, writers typically follow the ``Lasswell Communication Model'' theory  \cite{lasswell1948structure} (i.e., ``5W1H'' paradigm\footnote{The 5W1H paradigm: What, Why, Who, Where, When, and How. What and how can be packaged as the event. Who stands for entities, which are the participants in the event. Where stands for the location where the event occurred. When stands for date. Why stands for the result of the event.}). It includes five core news elements: \emph{event}, \emph{entity}, \emph{location}, \emph{date}, \emph{result}. It is essential to identify and extract these elements related to the topic within the news, since they constitute the factual backbone of the subsequent summaries. Therefore, we design a prompt consisting of five targeted guiding questions for each element, guiding the LLM in generating content faithful to the original news content: (1) What is the event happening in this document related to the keyword? (2) What are the important entities of this event in this document related to the keyword? (3) Where does the event happen? (4) What is the important date of this event in this document related to the keyword? (5) What is the result of this?
Additionally, we further incorporate a one-shot example into the prompt to help LLMs better extract elements. By providing the prompt, we can leverage the in-context learning capability of LLMs, enabling them to adapt their generation policy by conditioning on the given information.

\subsubsection{Element-based Summarization}
The second step is generating the summary. To balance faithfulness and completeness, we design a structured prompt that integrates the original article, the extracted elements, and a one-shot example. This guides the LLM to focus on topic-relevant facts while preserving necessary context, making it more likely to retain these details in the summary rather than redundant or less significant ones.


Through the multi-step reasoning process of the \emph{Element-CoT}, the extracted elements are integrated into the prompt, guaranteeing that the LLM has a clear reference to the most crucial details. Finally, we output a news summary in a specific format, such as: ``\textit{2005-01-03: President George W. Bush appoints former Presidents George H.W. Bush and Bill Clinton to head fundraising efforts for humanitarian aid after a devastating tsunami in Indian Ocean.}'' This approach helps maintain consistency between the generated content and the context, effectively reducing the risk of ambiguities and hallucinations in news summarization. All prompts in the \emph{Element-CoT} can be viewed in Figure~\ref{fig:prompts4Element-CoT}. 

\subsection{Date Selection}
\label{sub:ds}
To identify $l$ representative dates for constructing the timeline, we expect to measure whether a date is important through multiple dimensions. Our date selection module constructs an associated date graph to analyze temporal saliency and assess event prominence through event clustering. Details are demonstrated in Algorithm~\ref{alg_date}.

\subsubsection{Associated Date Graph Construction}
Considering that an event may not attract much attention at the time it occurs but gradually gains significance over time and is mentioned more frequently, we need to take its temporal saliency into account. We construct an associated date graph  $G = (V, E)$, where dates are linked based on their mention frequency, facilitating the analysis of temporal saliency. Dates with higher mention frequencies are considered to have greater temporal saliency. In this graph, each node $v_{t_i}$ represents a date $t_i$ mentioned in both the news articles $A$ and news summaries $\hat{S}$, and each edge represents a mention between the associated date nodes. For a given news article ${a_i}$ and its summary $\hat{s}_{a_i}$, edges are established in the following cases: (i) between the node $v_{t_p}$ of publication date ${t_p}$ and itself; (ii) between $v_{t_p}$ and $v_{t_d}$, where $t_d$ is the date mentioned in summary $\hat{s}_{a_i}$; (iii) between $v_{t_p}$ and $v_{t_s}$, where ${t_s}$ is the date mentioned in sentences containing the topic query keyphrases $Q$ in ${a_i}$. The temporal saliency of each date $t_i$ can be calculated by the in-degree $d_{in}(v_{t_i})$ in $G$.

\subsubsection{Event Clustering}
\label{sec:clustering}
As there are many similar news summaries, we aim to reduce redundancy while preserving subtle differences in TLS.
Inspired by \cite{hu-etal-2024-moments}, we adopt an event clustering method to identify and organize key events for timeline construction by grouping similar events together to determine whether a particular category of events is significant. Specifically, for each incoming news summary $\hat{s}_{a_i}$, we first retrieve similar news summaries based on vector similarity. Then, we use an LLM with few-shot prompting to perform pairwise event judgments between $\hat{s}_{a_i}$ and each retrieved summary $\hat{s}_{a_j}$. If $\hat{s}_{a_i}$ and $\hat{s}_{a_j}$ describe the same event, they are grouped into the same cluster. This process yields the date-event clusters $C$. 
Each cluster $C_k$ comprises a subset of event summaries. 
We select the most frequent date in cluster $C_k$ as its representative date, which can be calculated by:
 \begin{equation}
     t_c=\underset{t\in T\left(C_k \right)}{\mathop{\arg \max }}\,cnt\left( t, C_k \right),
 \end{equation}
where $T\left( C_k \right)$ represents the set of all dates associated with the event summaries in $C_k$, and $cnt$ is a counting function. The size of a cluster $C_k$, denoted by $|C_k|$, is the number of news summaries it contains. A larger $|C_k|$ indicates greater event prominence.

\subsubsection{Date Ranking}
Different from the previous date ranking method \cite{10.1145/2487788.2487829,gholipour-ghalandari-ifrim-2020-examining} that solely on the frequency of time mentions as a measure of temporal saliency, we consider both the date and event importance while ranking by calculating $score(t_i)$ of each date $t_i$ in the associated date graph $G$ and date-event clusters $C$:
\begin{equation}
score(t_i)=\alpha \cdot d_{in}\left(v_{t_i} \right)+\left( 1-\alpha  \right)\cdot \left| C\left(t_i \right) \right|
\end{equation}
where $\alpha$ serves as a hyperparameter to balance the weight between the temporal saliency and event prominence. We rank the date $t_i$ based on $score(t_i)$ and select the top $l$ dates with the highest scores to construct a timeline.

\subsection{Date-Event Summarization}
Each date on the timeline corresponds to an event cluster containing one or multiple news summaries with varying descriptions. In this section, we perform multi-document summarization and propose a \emph{Causal-CoT} approach. This approach can capture and leverage the causal relationship among multiple event summaries in the clusters, and finally help the LLM generate a complete, coherent, and concise timeline summary for a given topic. The intuition is that events within the same cluster have potential causal dependencies. This process can prevent the omission of critical context and enhance factual faithfulness.  

\subsubsection{Chunking}
An event cluster may contain more than 100 summaries, potentially exceeding the context length limit of the LLM input.
Moreover, longer inputs are more likely to induce hallucinations \cite{li2024improving}. To address this issue, we recursively divide each cluster $C_k$ into smaller chunks of maximum size $M_B$ and summarize each chunk, producing intermediate summaries $\hat{S}_{B}$. If the number of intermediate summaries exceeds $M_B$, they are further partitioned and summarized, repeating this process until all chunks are within the size limit, yielding the final date-event summary.

\subsubsection{Causal-CoT}
After chunking, we design a \emph{Causal-CoT} module to generate date-event summaries by explicitly eliciting causal dependencies across news within the same event cluster through CoT. \emph{Causal-CoT} consists of three steps: preliminary summarization, causal relationship analysis, and post-editing.

\paragraph{Preliminary Summarization.}
News elements analysis is indispensable for causal relationships identification. Similar to news summarization in Section~\ref{sub:ns}, we employ \emph{Element-CoT} to extract key news elements and generate preliminary summaries. Since the date of each event cluster has already been determined in the section~\ref{sec:clustering}, we mainly focus on events, entities, locations, and the results of news summaries in each event cluster. The output is the key elements for each event cluster chunk and its summary.

\paragraph{Causal Relationship Analysis.}
Within each event cluster, different news summaries often contain overlapping elements while also complementing one another. To understand the causal relationships across these summaries, we guide the LLM to recognize same elements across multiple news summaries and infer how one influences another. The model is then prompted to integrate these causal relationships into a coherent and chronologically ordered narrative, eliminating redundancy and constructing a logical chain of events from the dispersed descriptions.
Formally, given the event cluster $C(t_i)$ with extracted elements $\mathbb{E}=\left\{ {{\varepsilon }_{1}},{{\varepsilon }_{2}},\ldots ,{{\varepsilon }_{n}} \right\}$, the model outputs relations of the form $\left( {{{\hat{s}}}_{{{a}_{i}}}}\to {{{\hat{s}}}_{{{a}_{j}}}}:{{\varepsilon }_{1}}\to {{\varepsilon }_{2}},{{\varepsilon }_{3}}\to {{\varepsilon }_{4}} \right)$, where $\hat{s}_{a_i}$, $\hat{s}_{a_j}$ are news summaries. 
This prompt-based causal reasoning yields an interpretable list of causal links, which is later used for post-editing to enhance coherence and reduce redundancy.

\paragraph{Post-Editing.}
Based on the causal relationship list, we refine the preliminary summaries through post-editing. Specifically, the LLM consolidates overlapping elements and enriches the preliminary summary with causally related information in a coherent sequence, minimizing redundancy and omissions. 
For example, through causal relationship analysis, we obtain ``\textit{Medical treatment causes Saleh’s arrival in the United States.}'' This allows us to perform post-editing on the preliminary summary ``\textit{Saleh arrives in the United States.}'' and supplement it to become ``\textit{Saleh arrives in the United States for medical treatment.}'' All the prompts of \emph{Causal-CoT} used are provided in Figure~\ref{fig:prompts4Causal-CoT}.

\section{Experiments}
In this section, we evaluate \texttt{NTS-CoT} through experiments aimed at answering the following questions:
\textbf{RQ1.} How does \texttt{NTS-CoT} perform compared to the state-of-the-art baselines? \textbf{RQ2.} Does each module of \texttt{NTS-CoT} work effectively, and how do they impact the performance? \textbf{RQ3.} How do model parameters affect the method? \textbf{RQ4.} How do humans evaluate the generated TLS?

\subsection{Experimental Setup}
\textbf{Datasets.}
We perform experiments on three publicly available TLS datasets: Timeline17 (T17) \cite{tran2013leveraging}, Crisis \cite{tran2015timeline}, and Entities \cite{gholipour-ghalandari-ifrim-2020-examining}. Each dataset has different topics and multiple timelines annotated by experts. To provide more accurate and rich time information, we preprocess each news article using HeidelTime \cite{strotgen2013multilingual}, converting the time descriptions in the article (e.g., ``yesterday'', ``April 20'') into standard time stamps ``YYYY-MM-DD''.

\textbf{Baselines.}
We compare the performance of \emph{NTS-CoT} against four categories of baselines.
(i) \textit{Extractive methods.} These methods construct timelines by selecting salient sentences or events from source documents, including MARTSCHAT \cite{martschat-markert-2018-temporally}, DATEWISE \cite{gholipour-ghalandari-ifrim-2020-examining}, CLUST \cite{gholipour-ghalandari-ifrim-2020-examining}, SDF \cite{la2021summarize}, and EGC \cite{li-etal-2021-timeline}.
(ii) \textit{Abstractive method.} We compare with STEEN \cite{steen-markert-2019-abstractive}, which utilizes an unsupervised abstractive TLS system to generate readable summaries.
(iii) \textit{LLM-based method.} We compare with LLM-TLS \cite{hu-etal-2024-moments}, which leverages LLMs for news summarization and event clustering.
(iv) \textit{LLM + different prompting strategies.} Due to the limited availability of open-source LLM-based methods, we also compare with Zero-shot, One-shot, and Standard-CoT.
We employ Llama3-8B-Instruct \cite{dubey2024llama} as the foundational LLM.

\begin{table}[h]
\vspace{-0.5cm}
  \centering
  \caption{The comparison results of \texttt{NTS-CoT} and baselines on three TLS datasets. The best results are \textbf{bolded}, and the best results on LLM-based models are \underline{underlined}.}
  \resizebox{\textwidth}{!}{
    \begin{tabular}{lccccccccc}
    \hline
    \multicolumn{1}{c}{\multirow{2}[2]{*}{\textbf{Method}}} & \multicolumn{3}{c}{\textbf{T17}} & \multicolumn{3}{c}{\textbf{Crisis}} & \multicolumn{3}{c}{\textbf{Entities}} \bigstrut[t]\\
          & \textbf{AR-1} & \textbf{AR-2} & \textbf{Date F1} & \textbf{AR-1} & \textbf{AR-2} & \textbf{Date F1} & \textbf{AR-1} & \textbf{AR-2} & \textbf{Date F1} \bigstrut[b]\\
    \hline
    MARTSCHAT\cite{martschat-markert-2018-temporally} & 0.105  & 0.030  & 0.544  & 0.075  & 0.016  & 0.281  & 0.042  & 0.009  & 0.167  \bigstrut[t]\\
    DATEWISE\cite{gholipour-ghalandari-ifrim-2020-examining} & \textbf{0.120 } & \textbf{0.035 } & 0.544  & 0.089  & 0.026  & 0.295  & 0.057  & 0.017  & 0.205  \\
    CLUST\cite{gholipour-ghalandari-ifrim-2020-examining} & 0.082  & 0.020  & 0.407  & 0.061  & 0.013  & 0.226  & 0.051  & 0.015  & 0.174  \\
    SDF\cite{la2021summarize}   & 0.120  & 0.035  & \textbf{0.553 } & 0.086  & 0.018  & 0.302  & 0.051  & 0.014  & 0.197  \\
    EGC\cite{li-etal-2021-timeline}   & 0.103  & 0.024  & 0.550  & 0.079  & 0.015  & 0.291  & -     & -     & - \\
    STEEN\cite{steen-markert-2019-abstractive} & 0.093  & 0.024  & 0.512  & 0.082  & \textbf{0.027 } & 0.297  & -     & -     & - \bigstrut[b]\\
    \hline
    LLM-TLS\cite{hu-etal-2024-moments} & 0.094  & 0.024  & 0.512  & 0.089  & 0.022  & 0.288  & 0.075  & 0.027  & 0.209  \bigstrut[t]\\
    Zero-shot & 0.097  & 0.027  & 0.504  & 0.068  & 0.016  & 0.271  & 0.048  & 0.016  & 0.193  \\
    One-shot & 0.110  & 0.028  & 0.513  & 0.091  & 0.022  & 0.271  & 0.068  & 0.025  & 0.201  \\
    Standard-Cot & 0.089  & 0.025  & 0.521  & 0.059  & 0.014  & 0.257  & 0.050  & 0.015  & 0.214  \\
   \textbf{\textit{NTS-CoT (ours)}} & \underline{0.116}  & \underline{0.032}  & \underline{0.541}  & \textbf{\underline{0.094} } & \underline{0.024}  & \textbf{\underline{0.305} } & \textbf{\underline{0.075} } & \textbf{\underline{0.027} } & \textbf{\underline{0.230}}  \bigstrut[b]\\
    \hline
    \end{tabular}%
  }
  \label{tab:result}%
\end{table}%

\subsection{Overall Performance (RQ1)}
To validate the effectiveness of \texttt{NTS-CoT}, we perform a comprehensive comparison on the T17, Crisis, and Entities datasets. The main results are presented in Table~\ref{tab:result}, and we have several observations. 
\texttt{NTS-CoT} consistently outperforms all other LLM-based methods across the datasets. For instance, it achieves significant improvements of up to 23.4\% in AR-1, 33.3\% in AR-2, and 10.0\% in Date F1 compared to the LLM-TLS method.  These results indicate the ability of \texttt{NTS-CoT} to enhance the LLM's focus on key elements and relationships within news articles, effectively reducing hallucinations during the TLS task generation process, particularly in terms of accurate date identification.

Furthermore, \texttt{NTS-CoT} excels on the Crisis and Entities datasets, achieving the highest scores for AR-1 and Date F1 metrics. We find that \texttt{NTS-CoT} slightly trails behind the second-best non-LLM-based methods. The lower performance in AR metrics can be attributed to the LLM-based approach's tendency to introduce additional expressions and reorder words to improve summary coherence. Since the ground truth consists of sentences directly extracted from the news, non-LLM-based methods naturally achieve higher alignment scores. Despite this, the timeline summaries generated by \texttt{NTS-CoT} exhibit strong readability and coherence, demonstrating its practical utility. Moreover, our approach requires no additional training, offering lower costs and greater flexibility.

\subsection{Ablation Study (RQ2)}
We perform an ablation study on \texttt{NTS-CoT} by sequentially excluding both major components (\emph{Element-CoT} and \emph{Causal-CoT}) and the fine-grained sub-components within them. As shown in Table~\ref{tab:ablation2}, the results highlight the significance of each component.

At the component level, we replace \emph{Element-CoT} and \emph{Causal-CoT} with a straightforward instruction that generates one-sentence summaries.
Specifically, \texttt{NTS-CoT} consistently outperforms its variants across all datasets and evaluation metrics. Removing either component leads to declines in performance across all metrics and an increase in hallucinations within the TLS task. Notably, the most substantial performance drop occurs when the \emph{Element-CoT} is excluded, indicating its critical role in ensuring accurate date identification. This suggests that the \emph{Element-CoT} significantly enhances the LLM's ability to focus on key details in news content. Furthermore, removing the \emph{Causal CoT} results in a comparatively smaller reduction in AR scores. This can be attributed to the fact that the \emph{Causal CoT} has contained an initial summarization step followed by causal relationship analysis and post-editing. Given that news summaries within a cluster are already concise and contain overlapping information, the causal relationship analysis introduces only marginal additional content or modifications. As a result, the impact on textual overlap calculations remains relatively limited.

\begin{table*}[h]
  \centering
  \caption{Fine-grained ablation study.}
    \resizebox{\columnwidth}{!}{
    \begin{tabular}{lccccccccc}
    \hline
          & \multicolumn{3}{c}{\textbf{T17}} & \multicolumn{3}{c}{\textbf{Crisis}} & \multicolumn{3}{c}{\textbf{Entities}} \\
          & \textbf{AR-1} & \textbf{AR-2} & \textbf{Date F1} & \textbf{AR-1} & \textbf{AR-2} & \textbf{Date F1} & \textbf{AR-1} & \textbf{AR-2} & \textbf{Date F1} \\
    \hline
    {w/o event} & 0.105  & 0.026  & 0.523  & 0.092  & 0.022  & 0.305  & 0.069  & 0.026  & 0.231  \\
    {w/o entity} & 0.113  & 0.031  & 0.541  & 0.090  & 0.023  & 0.284  & 0.072  & 0.024  & 0.238  \\
    {w/o date} & 0.108  & 0.028  & 0.526  & 0.091  & 0.024  & 0.275  & 0.069  & 0.023  & 0.213  \\
    {w/o location} & 0.109  & 0.026  & 0.530  & 0.091  & 0.022  & 0.286  & 0.071  & 0.026  & 0.229  \\
    {w/o result} & 0.115  & 0.031  & 0.529  & 0.094  & 0.024  & 0.283  & 0.073  & 0.025  & 0.222  \\
    {w/o Element-CoT} & 0.092  & 0.022  & 0.521  & 0.088  & 0.023  & 0.284  & 0.074  & 0.026  & 0.212  \\
    \hline
    {w/o PS} & 0.115  & 0.031  & 0.541  & 0.093  & 0.023  & 0.305  & 0.073  & 0.026  & 0.230  \\
    {w/o CRA} & 0.115  & 0.031  & 0.541  & 0.092  & 0.023  & 0.305  & 0.073  & 0.026  & 0.230  \\
    {w/o Causal-CoT} & 0.115  & 0.031  & 0.541  & 0.091  & 0.023  & 0.305  & 0.073  & 0.026  & 0.230  \\
    \hline
    \textbf{\emph{NTS-CoT}} & \textbf{0.116}  & \textbf{0.032}  & \textbf{0.541}  & \textbf{0.094}  & \textbf{0.024}  & \textbf{0.305}  & \textbf{0.075}  & \textbf{0.027}  & \textbf{0.230}  \\
    \hline
    \end{tabular}
    }
  \label{tab:ablation2}%
\end{table*}%

At the sub-component level, we perform fine-grained ablations on Element-CoT and Causal-CoT. 
For Element-CoT, we sequentially remove each element, including event, entity, location, date, and result, and observe performance declines in all cases. Specifically, removing date leads to a significant drop in Date F1, highlighting the crucial role of date elements in maintaining temporal accuracy. Additionally, the event element contributes most to summary accuracy, while removing result causes a relatively smaller decline, possibly because the extracted result information overlaps with the other four elements to some extent.
For Causal-CoT, removing either preliminary summarization (w/o PS) or causal reasoning (w/o CRA) also degrades performance, indicating that both contribute to improving coherence and mitigating hallucinations in date-event summarization.


\subsection{Parameter Analysis (RQ3)}

\begin{figure*}[t]
    \centering
    
    \begin{subfigure}[t]{0.32\textwidth}
        \centering
        \resizebox{\linewidth}{3cm}{
            \includegraphics{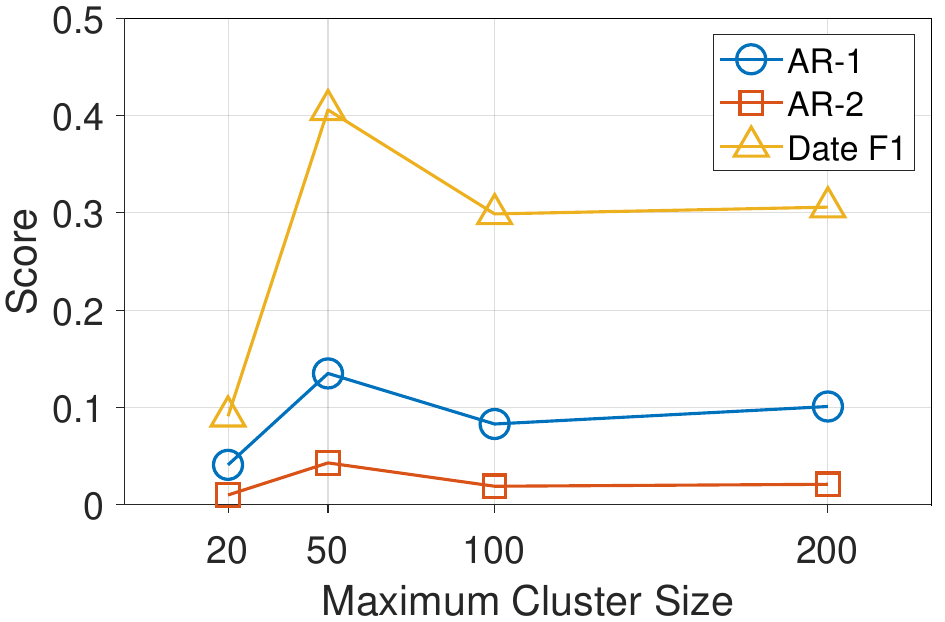}
        }
        \caption{Impact of maximum cluster sizes on Crisis.}
        \label{fig:max_cluster_size_a}
    \end{subfigure}
    \hfill
    \begin{subfigure}[t]{0.32\textwidth}
        \centering
        \resizebox{\linewidth}{3cm}{
            \includegraphics{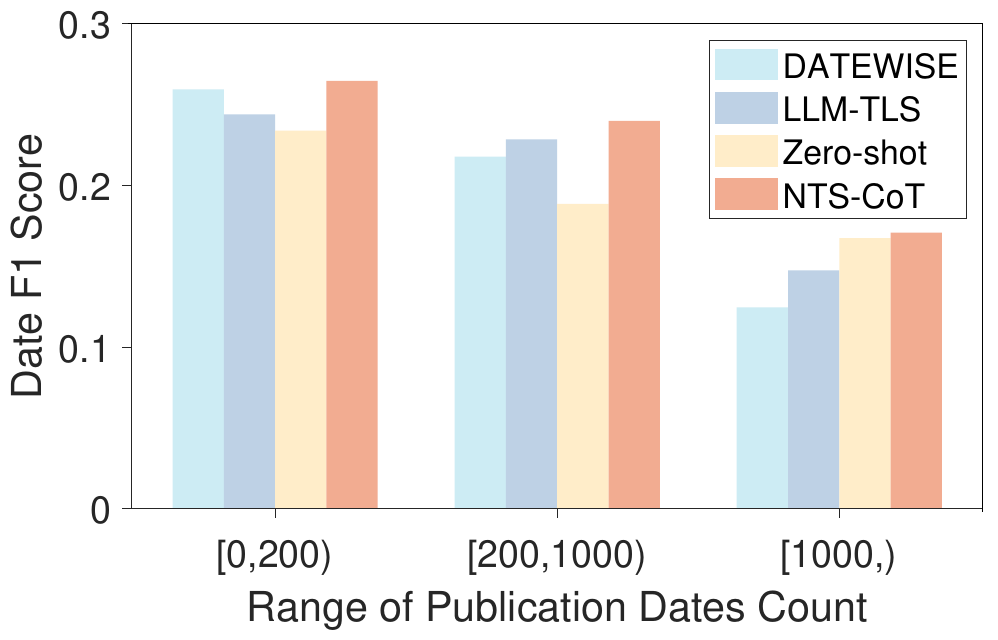}
        }
        \caption{Impact of the number of publication dates on Entities.}
        \label{fig:pub_date_b}
    \end{subfigure}
    \hfill
    \begin{subfigure}[t]{0.32\textwidth}
        \centering
        \resizebox{\linewidth}{3cm}{
            \includegraphics{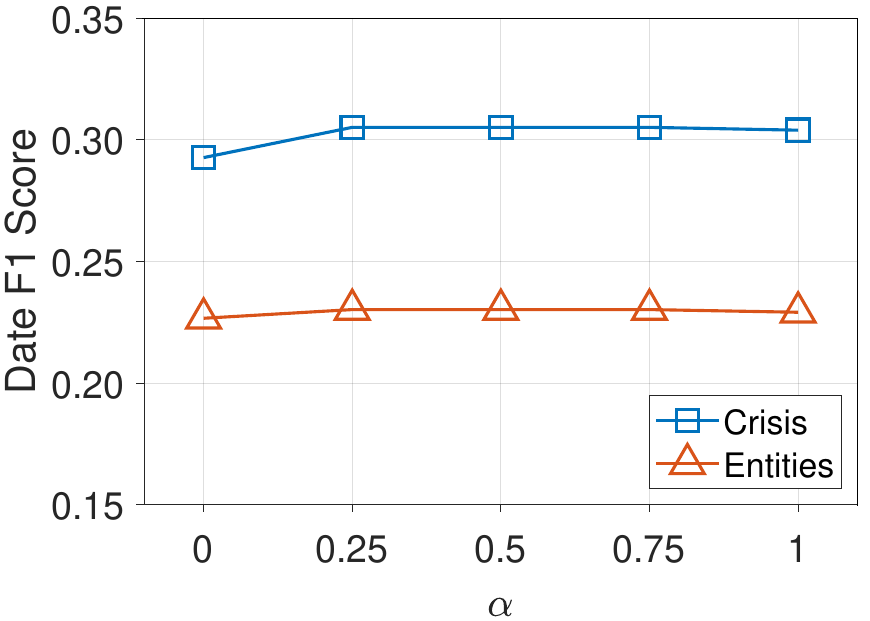}
        }
        \caption{Impact of the $\alpha$ in Date ranking on Crisis and Entities dataset.}
        \label{fig:alpha_c}
    \end{subfigure}
    
    \caption{Parameter analysis results.}
    \label{fig:parameter_analysis}
    \vspace{-0.5cm}
\end{figure*}

\paragraph{Impact of maximum cluster size.} 
In the date-event summarization stage, each cluster may contain multiple news summaries. A larger cluster size increases the number of news summaries and the complexity of relationships among them, which raises the likelihood of hallucinations when the LLM generates date-event summaries, ultimately degrading summarization performance. To address this challenge, \emph{Causal CoT} is specifically designed to mitigate such issues. We investigate the impact of cluster size on summarization performance using the Crisis dataset, as it features a higher average number of news articles per topic compared to other datasets. The results in Figure~\ref{fig:parameter_analysis} (a) demonstrate that performance remains consistently stable even when the cluster size reaches 200. This highlights the effectiveness of \emph{Causal CoT} in managing complex multi-document relationships and maintaining robust summarization quality.

\paragraph{Impact of the number of publication dates.} 
We investigate the impact of the number of publication dates on the temporal accuracy of timeline construction. As the number of publication dates increases, the complexity of selecting milestone nodes also rises, resulting in a corresponding decrease in performance. As illustrated in Figure~\ref{fig:parameter_analysis} (b), \texttt{NTS-CoT} consistently outperforms both DATEWISE and LLM-TLS as the number of publication dates grows, highlighting its effectiveness in mitigating hallucinations in LLM-generated time descriptions. Furthermore, the results indicate that \texttt{NTS-CoT} exhibits robust capability in handling timeline summarization tasks across varying date spans, underscoring its adaptability and reliability.

\paragraph{Impact of weight $\alpha$.}
The $\alpha$ parameter is used to balance the weight between temporal saliency and event prominence in date ranking. We gradually increase $\alpha$ from 0 to 1 to observe its impact on model performance. As shown in Figure~\ref{fig:parameter_analysis} (c), the results indicate that when $\alpha=1$ (excluding date importance from the associated date graph) or $\alpha=0$ (excluding event importance from event clustering), there is a slight decline in performance. However, for other values of $\alpha$, the performance remains stable. 
In most cases, the associated date graph and event clustering highlight similar high-importance dates.
Their integration can achieve complementarity, leading to a more accurate date ranking. Given this observation, we finally selected a balanced middle value of $\alpha$, which proved to be a relatively stable choice across datasets.

\begin{wraptable}{r}{0.45\textwidth}
\vspace{-5em}
  \centering
  \caption{The percentage of human preferences for NTS-CoT and LLM-TLS on the Faithfulness and Completeness metrics.}
  \resizebox{0.45\textwidth}{!}{
    \begin{tabular}{l|cccc}
    \hline
          & \multicolumn{2}{c}{\textbf{Faithfulness}} & \multicolumn{2}{c}{\textbf{Completeness}} \\
          & \texttt{\textbf{NTS-CoT}} & LLM-TLS & \texttt{\textbf{NTS-CoT}} & LLM-TLS \\
    \hline
    \textbf{Preference} & 67.74\% & 32.26\% & 54.38\% & 45.62\% \\
    \hline
    \end{tabular}%
  \label{tab:human}%
  }
  \vspace{-0.3cm}
\end{wraptable}

\subsection{Human Evaluation (RQ4)}
To further evaluate our approach, we randomly select 30 timelines generated by LLM-TLS and \texttt{NTS-CoT} across three datasets for pairwise human evaluation on faithfulness and completeness, with detailed protocols and a case study provided in Appendix~\ref{sec:appendix_human}.
As shown in Table~\ref{tab:human}, among the 30 evaluated timelines, which contain 217 date-event summaries with identical dates, \texttt{NTS-CoT} is preferred 67.74\% of the time for Faithfulness and 54.38\% for Completeness. These results further validate the effectiveness of \texttt{NTS-CoT} in reducing hallucinations. 

\section{Related Works}
\subsection{News Timeline Summarization}
Timeline Summarization (TLS) aims to generate summaries of significant events from multiple documents \cite{swan2000automatic} and has been extensively explored in previous research. Existing research has mainly included extractive and abstractive methods. Extractive methods mainly involve sentence-level extraction, selecting representative sentences from multiple documents to construct the timeline \cite{mansouri2023towards,chieu2004query,yan2011timeline,10.1145/2487788.2487829,martschat-markert-2018-temporally,gholipour-ghalandari-ifrim-2020-examining}. In abstractive methods, Steen et al. \cite{steen-markert-2019-abstractive} proposed an unsupervised system to generate a summary for each cluster of the same event. Chen et al. \cite{chen2019learning} designed a memory-based TLS model, and improved it by incorporating the information interaction between events \cite{chen2023follow}. However, these methods are not aligned with human language habits due to semantic understanding and information integration limitations. 
Recent LLMs have shown significant potential in timeline summarization \cite{sojitra2024timeline,hu-etal-2024-moments}. Wang et al. \cite{wang2023web} leveraged extended task prompting to enhance news relevance identification. Song et al. \cite{song-etal-2024-combining} combined hierarchical VAEs with LLMs to generate summaries. Wu et al. \cite{wu2025unfolding} incorporate open-domain retrieval and introduce a new TLS dataset, while Qorib et al. \cite{qorib2025just} proposes constrained TLS. However, they mainly focus on defining new tasks and building datasets for LLMs in the TLS field to adapt to various tasks, but pay limited attention to improving the faithfulness and linguistic quality of the generated summaries. As a result, hallucination issues remain underexplored in LLM-based TLS.

\subsection{Hallucinations Mitigation in Summarization}
LLMs may generate results with hallucinations in summarization tasks. To address this issue, some existing studies focus on detecting or reducing hallucinations through pretraining~\cite{shen2023misleading}, fine-tuning~\cite{han-etal-2024-alignsum,wu2024alleviating,zhang2024r}, or reinforcement learning~\cite{roit-etal-2023-factually}. However, these methods are resource-intensive, as they typically require large-scale annotated datasets, significant computational resources, and extensive training time to adapt LLMs to specific domains or tasks. This process is often costly and inflexible, and limits adaptability to rapidly changing data.
Some studies have improved the faithfulness of generated summaries by modifying decoding strategies. \cite{van-der-poel-etal-2022-mutual} incorporate pointwise mutual information into beam search, while \cite{chae-etal-2024-mitigating} extend this with domain-conditional mutual information. \cite{shi-etal-2024-trusting,zhang2023alleviating} increase the probability of faithful content output through contrast decoding. However, modifying the decoding strategy is often unstable, as it may interfere with the model’s generation probabilities in unintended ways, potentially leading to incoherent or nonsensical sentences and even amplifying hallucinations.
In contrast, prompt-based methods offer a more lightweight and flexible alternative, enabling adaptation without modifying model parameters. Using prompts for consistency checks \cite{manakul-etal-2023-selfcheckgpt,li-etal-2024-self} or reflective post-editing \cite{zhao-etal-2023-verify} can also help mitigate hallucinations in summarization. Recently, Chain-of-thought (CoT) techniques have also been proven to be effective in summarization tasks \cite{he2022rethinking,wang-etal-2023-element,zhang20243a,gupta2025autosumm}. However, these methods are proposed for single-document summarization and rarely consider the complexity of multiple documents. In this work, we specifically target hallucinations in news timeline summarization across multiple documents.

\section{Conclusion}
In this work, we propose \texttt{NTS-CoT}, a novel framework that incorporates one date selection module and two CoT-based modules: \textit{Element-CoT} and \textit{Causal-CoT}. CoT-based modules are designed to mitigate hallucinations at two key stages of LLM-based news timeline summarization: unfaithful content during news summarization and information omission in date-event summarization. Experimental results demonstrate its effectiveness in alleviating hallucinations in TLS tasks. Our research reveals critical hallucination challenges in TLS and lays the groundwork for future exploration of mitigation strategies in LLM-based timeline summarization.



{
\small

\bibliographystyle{IEEEtran}
\bibliography{references}

\begin{thebibliography}{10}
\providecommand{\url}[1]{#1}
\csname url@samestyle\endcsname
\providecommand{\newblock}{\relax}
\providecommand{\bibinfo}[2]{#2}
\providecommand{\BIBentrySTDinterwordspacing}{\spaceskip=0pt\relax}
\providecommand{\BIBentryALTinterwordstretchfactor}{4}
\providecommand{\BIBentryALTinterwordspacing}{\spaceskip=\fontdimen2\font plus
\BIBentryALTinterwordstretchfactor\fontdimen3\font minus \fontdimen4\font\relax}
\providecommand{\BIBforeignlanguage}[2]{{%
\expandafter\ifx\csname l@#1\endcsname\relax
\typeout{** WARNING: IEEEtran.bst: No hyphenation pattern has been}%
\typeout{** loaded for the language `#1'. Using the pattern for}%
\typeout{** the default language instead.}%
\else
\language=\csname l@#1\endcsname
\fi
#2}}
\providecommand{\BIBdecl}{\relax}
\BIBdecl

\bibitem{10.1145/2487788.2487829}
G.~Binh~Tran, M.~Alrifai, and D.~Quoc~Nguyen, ``Predicting relevant news events for timeline summaries,'' in \emph{Proceedings of the 22nd International Conference on World Wide Web}, ser. WWW '13 Companion.\hskip 1em plus 0.5em minus 0.4em\relax New York, NY, USA: Association for Computing Machinery, 2013, p. 91–92.

\bibitem{tran2015joint}
G.~Tran, E.~Herder, and K.~Markert, ``Joint graphical models for date selection in timeline summarization,'' in \emph{Proceedings of the 53rd Annual Meeting of the Association for Computational Linguistics and the 7th International Joint Conference on Natural Language Processing (Volume 1: Long Papers)}.\hskip 1em plus 0.5em minus 0.4em\relax Beijing, China: Association for Computational Linguistics, Jul. 2015, pp. 1598--1607.

\bibitem{gholipour-ghalandari-ifrim-2020-examining}
D.~Gholipour~Ghalandari and G.~Ifrim, ``Examining the state-of-the-art in news timeline summarization,'' in \emph{Proceedings of the 58th Annual Meeting of the Association for Computational Linguistics}, D.~Jurafsky, J.~Chai, N.~Schluter, and J.~Tetreault, Eds.\hskip 1em plus 0.5em minus 0.4em\relax Online: Association for Computational Linguistics, Jul. 2020, pp. 1322--1334.

\bibitem{chieu2004query}
H.~L. Chieu and Y.~K. Lee, ``Query based event extraction along a timeline,'' in \emph{Proceedings of the 27th annual international ACM SIGIR conference on Research and development in information retrieval}, 2004, pp. 425--432.

\bibitem{martschat-markert-2018-temporally}
S.~Martschat and K.~Markert, ``A temporally sensitive submodularity framework for timeline summarization,'' in \emph{Proceedings of the 22nd Conference on Computational Natural Language Learning}, A.~Korhonen and I.~Titov, Eds.\hskip 1em plus 0.5em minus 0.4em\relax Brussels, Belgium: Association for Computational Linguistics, Oct. 2018, pp. 230--240.

\bibitem{brown2020language}
T.~Brown, B.~Mann, N.~Ryder, M.~Subbiah, J.~D. Kaplan, P.~Dhariwal, A.~Neelakantan, P.~Shyam, G.~Sastry, A.~Askell, S.~Agarwal, A.~Herbert-Voss, G.~Krueger, T.~Henighan, R.~Child, A.~Ramesh, D.~Ziegler, J.~Wu, C.~Winter, C.~Hesse, M.~Chen, E.~Sigler, M.~Litwin, S.~Gray, B.~Chess, J.~Clark, C.~Berner, S.~McCandlish, A.~Radford, I.~Sutskever, and D.~Amodei, ``Language models are few-shot learners,'' in \emph{Advances in Neural Information Processing Systems}, H.~Larochelle, M.~Ranzato, R.~Hadsell, M.~Balcan, and H.~Lin, Eds., vol.~33.\hskip 1em plus 0.5em minus 0.4em\relax Curran Associates, Inc., 2020, pp. 1877--1901.

\bibitem{zhang2023extractive}
H.~Zhang, X.~Liu, and J.~Zhang, ``Extractive summarization via chatgpt for faithful summary generation,'' in \emph{Findings of the association for computational linguistics: EMNLP 2023}, 2023, pp. 3270--3278.

\bibitem{ye2022unreliability}
X.~Ye and G.~Durrett, ``The unreliability of explanations in few-shot prompting for textual reasoning,'' in \emph{Advances in Neural Information Processing Systems}, S.~Koyejo, S.~Mohamed, A.~Agarwal, D.~Belgrave, K.~Cho, and A.~Oh, Eds., vol.~35.\hskip 1em plus 0.5em minus 0.4em\relax Curran Associates, Inc., 2022, pp. 30\,378--30\,392.

\bibitem{tam-etal-2023-evaluating}
D.~Tam, A.~Mascarenhas, S.~Zhang, S.~Kwan, M.~Bansal, and C.~Raffel, ``Evaluating the factual consistency of large language models through news summarization,'' in \emph{Findings of the Association for Computational Linguistics: ACL 2023}, A.~Rogers, J.~Boyd-Graber, and N.~Okazaki, Eds.\hskip 1em plus 0.5em minus 0.4em\relax Toronto, Canada: Association for Computational Linguistics, Jul. 2023, pp. 5220--5255.

\bibitem{ji2023survey}
Z.~Ji, N.~Lee, R.~Frieske, T.~Yu, D.~Su, Y.~Xu, E.~Ishii, Y.~J. Bang, A.~Madotto, and P.~Fung, ``Survey of hallucination in natural language generation,'' \emph{ACM Computing Surveys}, vol.~55, no.~12, pp. 1--38, 2023.

\bibitem{hase2023methods}
P.~Hase, M.~Diab, A.~Celikyilmaz, X.~Li, Z.~Kozareva, V.~Stoyanov, M.~Bansal, and S.~Iyer, ``Methods for measuring, updating, and visualizing factual beliefs in language models,'' in \emph{Proceedings of the 17th Conference of the European Chapter of the Association for Computational Linguistics}.\hskip 1em plus 0.5em minus 0.4em\relax Dubrovnik, Croatia: Association for Computational Linguistics, May 2023, pp. 2714--2731.

\bibitem{thakur2025assessing}
N.~Thakur, R.~Pradeep, S.~Upadhyay, D.~Campos, N.~Craswell, I.~Soboroff, H.~T. Dang, and J.~Lin, ``Assessing support for the trec 2024 rag track: A large-scale comparative study of llm and human evaluations,'' in \emph{Proceedings of the 48th International ACM SIGIR Conference on Research and Development in Information Retrieval}, 2025, pp. 2759--2763.

\bibitem{hu-etal-2024-moments}
Q.~Hu, G.~Moon, and H.~T. Ng, ``From moments to milestones: Incremental timeline summarization leveraging large language models,'' in \emph{Proceedings of the 62nd Annual Meeting of the Association for Computational Linguistics (Volume 1: Long Papers)}, L.-W. Ku, A.~Martins, and V.~Srikumar, Eds.\hskip 1em plus 0.5em minus 0.4em\relax Bangkok, Thailand: Association for Computational Linguistics, Aug. 2024, pp. 7232--7246.

\bibitem{wang2023web}
S.~Wang, Y.~Li, H.~Xiao, L.~Deng, and Y.~Dong, ``Web news timeline generation with extended task prompting,'' \emph{arXiv preprint arXiv:2311.11652}, 2023.

\bibitem{maynez-etal-2020-faithfulness}
J.~Maynez, S.~Narayan, B.~Bohnet, and R.~McDonald, ``On faithfulness and factuality in abstractive summarization,'' in \emph{Proceedings of the 58th Annual Meeting of the Association for Computational Linguistics}, D.~Jurafsky, J.~Chai, N.~Schluter, and J.~Tetreault, Eds.\hskip 1em plus 0.5em minus 0.4em\relax Online: Association for Computational Linguistics, Jul. 2020, pp. 1906--1919.

\bibitem{berezin2023named}
S.~Berezin and T.~Batura, ``Named entity inclusion in abstractive text summarization,'' in \emph{Proceedings of the Third Workshop on Scholarly Document Processing}, 2022, pp. 158--162.

\bibitem{tonmoy2024comprehensive}
S.~Tonmoy, S.~Zaman, V.~Jain, A.~Rani, V.~Rawte, A.~Chadha, and A.~Das, ``A comprehensive survey of hallucination mitigation techniques in large language models,'' \emph{arXiv preprint arXiv:2401.01313}, vol.~6, 2024.

\bibitem{huang2025survey}
L.~Huang, W.~Yu, W.~Ma, W.~Zhong, Z.~Feng, H.~Wang, Q.~Chen, W.~Peng, X.~Feng, B.~Qin \emph{et~al.}, ``A survey on hallucination in large language models: Principles, taxonomy, challenges, and open questions,'' \emph{ACM Transactions on Information Systems}, vol.~43, no.~2, pp. 1--55, 2025.

\bibitem{wei2022chain}
J.~Wei, X.~Wang, D.~Schuurmans, M.~Bosma, b.~ichter, F.~Xia, E.~Chi, Q.~V. Le, and D.~Zhou, ``Chain-of-thought prompting elicits reasoning in large language models,'' in \emph{Advances in Neural Information Processing Systems}, S.~Koyejo, S.~Mohamed, A.~Agarwal, D.~Belgrave, K.~Cho, and A.~Oh, Eds., vol.~35.\hskip 1em plus 0.5em minus 0.4em\relax Curran Associates, Inc., 2022, pp. 24\,824--24\,837.

\bibitem{yao2023tree}
S.~Yao, D.~Yu, J.~Zhao, I.~Shafran, T.~Griffiths, Y.~Cao, and K.~Narasimhan, ``Tree of thoughts: Deliberate problem solving with large language models,'' in \emph{Advances in Neural Information Processing Systems}, A.~Oh, T.~Naumann, A.~Globerson, K.~Saenko, M.~Hardt, and S.~Levine, Eds., vol.~36.\hskip 1em plus 0.5em minus 0.4em\relax Curran Associates, Inc., 2023, pp. 11\,809--11\,822.

\bibitem{wang-etal-2023-element}
Y.~Wang, Z.~Zhang, and R.~Wang, ``Element-aware summarization with large language models: Expert-aligned evaluation and chain-of-thought method,'' in \emph{Proceedings of the 61st Annual Meeting of the Association for Computational Linguistics (Volume 1: Long Papers)}, A.~Rogers, J.~Boyd-Graber, and N.~Okazaki, Eds.\hskip 1em plus 0.5em minus 0.4em\relax Toronto, Canada: Association for Computational Linguistics, Jul. 2023, pp. 8640--8665.

\bibitem{tran2013leveraging}
G.~B. Tran, T.~A. Tran, N.-K. Tran, M.~Alrifai, and N.~Kanhabua, ``Leveraging learning to rank in an optimization framework for timeline summarization,'' in \emph{SIGIR 2013 Workshop on Time-aware Information Access (TAIA}, 2013.

\bibitem{tran2015timeline}
G.~Tran, M.~Alrifai, and E.~Herder, ``Timeline summarization from relevant headlines,'' in \emph{Advances in Information Retrieval: 37th European Conference on IR Research, ECIR 2015, Vienna, Austria, March 29-April 2, 2015. Proceedings 37}.\hskip 1em plus 0.5em minus 0.4em\relax Springer, 2015, pp. 245--256.

\bibitem{lasswell1948structure}
H.~D. Lasswell, ``The structure and function of communication in society,'' \emph{The communication of ideas}, vol.~37, no.~1, pp. 136--139, 1948.

\bibitem{li2024improving}
T.~Li, Z.~Li, and Y.~Zhang, ``Improving faithfulness of large language models in summarization via sliding generation and self-consistency,'' in \emph{Proceedings of the 2024 Joint International Conference on Computational Linguistics, Language Resources and Evaluation (LREC-COLING 2024)}, 2024, pp. 8804--8817.

\bibitem{strotgen2013multilingual}
J.~Str{\"o}tgen and M.~Gertz, ``Multilingual and cross-domain temporal tagging,'' \emph{Language Resources and Evaluation}, vol.~47, pp. 269--298, 2013.

\bibitem{la2021summarize}
M.~La~Quatra, L.~Cagliero, E.~Baralis, A.~Messina, and M.~Montagnuolo, ``Summarize dates first: A paradigm shift in timeline summarization,'' in \emph{Proceedings of the 44th International ACM SIGIR Conference on Research and Development in Information Retrieval}, 2021, pp. 418--427.

\bibitem{li-etal-2021-timeline}
M.~Li, T.~Ma, M.~Yu, L.~Wu, T.~Gao, H.~Ji, and K.~McKeown, ``Timeline summarization based on event graph compression via time-aware optimal transport,'' in \emph{Proceedings of the 2021 Conference on Empirical Methods in Natural Language Processing}, M.-F. Moens, X.~Huang, L.~Specia, and S.~W.-t. Yih, Eds.\hskip 1em plus 0.5em minus 0.4em\relax Online and Punta Cana, Dominican Republic: Association for Computational Linguistics, Nov. 2021, pp. 6443--6456.

\bibitem{steen-markert-2019-abstractive}
J.~Steen and K.~Markert, ``Abstractive timeline summarization,'' in \emph{Proceedings of the 2nd Workshop on New Frontiers in Summarization}, L.~Wang, J.~C.~K. Cheung, G.~Carenini, and F.~Liu, Eds.\hskip 1em plus 0.5em minus 0.4em\relax Hong Kong, China: Association for Computational Linguistics, Nov. 2019, pp. 21--31.

\bibitem{dubey2024llama}
A.~Dubey, A.~Jauhri, A.~Pandey, A.~Kadian, A.~Al-Dahle, A.~Letman, A.~Mathur, A.~Schelten, A.~Yang, A.~Fan \emph{et~al.}, ``The llama 3 herd of models,'' \emph{arXiv preprint arXiv:2407.21783}, 2024.

\bibitem{swan2000automatic}
R.~Swan and J.~Allan, ``Automatic generation of overview timelines,'' in \emph{Proceedings of the 23rd annual international ACM SIGIR conference on Research and development in information retrieval}, 2000, pp. 49--56.

\bibitem{mansouri2023towards}
B.~Mansouri, R.~Campos, and A.~Jatowt, ``Towards timeline generation with abstract meaning representation,'' in \emph{Companion Proceedings of the ACM Web Conference 2023}, 2023, pp. 1204--1207.

\bibitem{yan2011timeline}
R.~Yan, L.~Kong, C.~Huang, X.~Wan, X.~Li, and Y.~Zhang, ``Timeline generation through evolutionary trans-temporal summarization,'' in \emph{Proceedings of the 2011 Conference on Empirical Methods in Natural Language Processing}, 2011, pp. 433--443.

\bibitem{chen2019learning}
X.~Chen, Z.~Chan, S.~Gao, M.-H. Yu, D.~Zhao, and R.~Yan, ``Learning towards abstractive timeline summarization,'' in \emph{Proceedings of the Twenty-Eighth International Joint Conference on Artificial Intelligence, {IJCAI-19}}.\hskip 1em plus 0.5em minus 0.4em\relax International Joint Conferences on Artificial Intelligence Organization, 7 2019, pp. 4939--4945.

\bibitem{chen2023follow}
X.~Chen, M.~Li, S.~Gao, Z.~Chan, D.~Zhao, X.~Gao, X.~Zhang, and R.~Yan, ``Follow the timeline! generating an abstractive and extractive timeline summary in chronological order,'' \emph{ACM Transactions on Information Systems}, vol.~41, no.~1, pp. 1--30, 2023.

\bibitem{sojitra2024timeline}
D.~Sojitra, R.~Jain, S.~Saha, A.~Jatowt, and M.~Gupta, ``Timeline summarization in the era of llms,'' in \emph{Proceedings of the 47th International ACM SIGIR Conference on Research and Development in Information Retrieval}, 2024, pp. 2657--2661.

\bibitem{song-etal-2024-combining}
J.~Song, J.~Chim, A.~Tsakalidis, J.~Ive, D.~Atzil-Slonim, and M.~Liakata, ``Combining hierachical {VAE}s with {LLM}s for clinically meaningful timeline summarisation in social media,'' in \emph{Findings of the Association for Computational Linguistics: ACL 2024}, L.-W. Ku, A.~Martins, and V.~Srikumar, Eds.\hskip 1em plus 0.5em minus 0.4em\relax Bangkok, Thailand: Association for Computational Linguistics, Aug. 2024, pp. 14\,651--14\,672.

\bibitem{wu2025unfolding}
W.~Wu, S.~Huang, Y.~Jiang, P.~Xie, F.~Huang, and H.~Zhao, ``Unfolding the headline: Iterative self-questioning for news retrieval and timeline summarization,'' in \emph{Findings of the Association for Computational Linguistics: NAACL 2025}, 2025, pp. 4385--4398.

\bibitem{qorib2025just}
M.~R. Qorib, Q.~Hu, and H.~T. Ng, ``Just what you desire: Constrained timeline summarization with self-reflection for enhanced relevance,'' in \emph{Proceedings of the AAAI Conference on Artificial Intelligence}, vol.~39, no.~23, 2025, pp. 25\,065--25\,073.

\bibitem{shen2023misleading}
J.~Shen, J.~Liu, D.~Finnie, N.~Rahmati, M.~Bendersky, and M.~Najork, ``“why is this misleading?”: Detecting news headline hallucinations with explanations,'' in \emph{Proceedings of the ACM Web Conference 2023}, 2023, pp. 1662--1672.

\bibitem{han-etal-2024-alignsum}
Y.~Han, Y.~Wang, R.~Wang, L.~Chen, and K.~Yu, ``{A}lign{S}um: Data pyramid hierarchical fine-tuning for aligning with human summarization preference,'' in \emph{Findings of the Association for Computational Linguistics: EMNLP 2024}, Y.~Al-Onaizan, M.~Bansal, and Y.-N. Chen, Eds.\hskip 1em plus 0.5em minus 0.4em\relax Miami, Florida, USA: Association for Computational Linguistics, Nov. 2024, pp. 8506--8522.

\bibitem{wu2024alleviating}
Y.~Wu, Y.~Wang, T.~Chen, C.~Liu, N.~Xi, Q.~Gu, H.~Lei, Z.~Jiang, Y.~Chen, and L.~Ji, ``Alleviating hallucinations in large language models with scepticism modeling,'' \emph{arXiv preprint arXiv:2409.06601}, 2024.

\bibitem{zhang2024r}
H.~Zhang, S.~Diao, Y.~Lin, Y.~Fung, Q.~Lian, X.~Wang, Y.~Chen, H.~Ji, and T.~Zhang, ``R-tuning: Instructing large language models to say ‘i don’t know’,'' in \emph{Proceedings of the 2024 Conference of the North American Chapter of the Association for Computational Linguistics: Human Language Technologies (Volume 1: Long Papers)}.\hskip 1em plus 0.5em minus 0.4em\relax Mexico City, Mexico: Association for Computational Linguistics, Jun. 2024, pp. 7106--7132.

\bibitem{roit-etal-2023-factually}
P.~Roit, J.~Ferret, L.~Shani, R.~Aharoni, G.~Cideron, R.~Dadashi, M.~Geist, S.~Girgin, L.~Hussenot, O.~Keller, N.~Momchev, S.~Ramos~Garea, P.~Stanczyk, N.~Vieillard, O.~Bachem, G.~Elidan, A.~Hassidim, O.~Pietquin, and I.~Szpektor, ``Factually consistent summarization via reinforcement learning with textual entailment feedback,'' in \emph{Proceedings of the 61st Annual Meeting of the Association for Computational Linguistics (Volume 1: Long Papers)}, A.~Rogers, J.~Boyd-Graber, and N.~Okazaki, Eds.\hskip 1em plus 0.5em minus 0.4em\relax Toronto, Canada: Association for Computational Linguistics, Jul. 2023, pp. 6252--6272.

\bibitem{van-der-poel-etal-2022-mutual}
L.~van~der Poel, R.~Cotterell, and C.~Meister, ``Mutual information alleviates hallucinations in abstractive summarization,'' in \emph{Proceedings of the 2022 Conference on Empirical Methods in Natural Language Processing}, Y.~Goldberg, Z.~Kozareva, and Y.~Zhang, Eds.\hskip 1em plus 0.5em minus 0.4em\relax Abu Dhabi, United Arab Emirates: Association for Computational Linguistics, Dec. 2022, pp. 5956--5965.

\bibitem{chae-etal-2024-mitigating}
K.~Chae, J.~Choi, Y.~Jo, and T.~Kim, ``Mitigating hallucination in abstractive summarization with domain-conditional mutual information,'' in \emph{Findings of the Association for Computational Linguistics: NAACL 2024}, K.~Duh, H.~Gomez, and S.~Bethard, Eds.\hskip 1em plus 0.5em minus 0.4em\relax Mexico City, Mexico: Association for Computational Linguistics, Jun. 2024, pp. 1809--1820.

\bibitem{shi-etal-2024-trusting}
W.~Shi, X.~Han, M.~Lewis, Y.~Tsvetkov, L.~Zettlemoyer, and W.-t. Yih, ``Trusting your evidence: Hallucinate less with context-aware decoding,'' in \emph{Proceedings of the 2024 Conference of the North American Chapter of the Association for Computational Linguistics: Human Language Technologies (Volume 2: Short Papers)}, K.~Duh, H.~Gomez, and S.~Bethard, Eds.\hskip 1em plus 0.5em minus 0.4em\relax Mexico City, Mexico: Association for Computational Linguistics, Jun. 2024, pp. 783--791.

\bibitem{zhang2023alleviating}
Y.~Zhang, L.~Cui, S.~Shi \emph{et~al.}, ``Alleviating hallucinations of large language models through induced hallucinations,'' in \emph{Findings of the Association for Computational Linguistics: NAACL 2025}, 2025, pp. 8218--8232.

\bibitem{manakul-etal-2023-selfcheckgpt}
P.~Manakul, A.~Liusie, and M.~Gales, ``{S}elf{C}heck{GPT}: Zero-resource black-box hallucination detection for generative large language models,'' in \emph{Proceedings of the 2023 Conference on Empirical Methods in Natural Language Processing}, H.~Bouamor, J.~Pino, and K.~Bali, Eds.\hskip 1em plus 0.5em minus 0.4em\relax Singapore: Association for Computational Linguistics, Dec. 2023, pp. 9004--9017.

\bibitem{li-etal-2024-self}
M.~Li, B.~Peng, M.~Galley, J.~Gao, and Z.~Zhang, ``Self-checker: Plug-and-play modules for fact-checking with large language models,'' in \emph{Findings of the Association for Computational Linguistics: NAACL 2024}, K.~Duh, H.~Gomez, and S.~Bethard, Eds.\hskip 1em plus 0.5em minus 0.4em\relax Mexico City, Mexico: Association for Computational Linguistics, Jun. 2024, pp. 163--181.

\bibitem{zhao-etal-2023-verify}
R.~Zhao, X.~Li, S.~Joty, C.~Qin, and L.~Bing, ``Verify-and-edit: A knowledge-enhanced chain-of-thought framework,'' in \emph{Proceedings of the 61st Annual Meeting of the Association for Computational Linguistics (Volume 1: Long Papers)}, A.~Rogers, J.~Boyd-Graber, and N.~Okazaki, Eds.\hskip 1em plus 0.5em minus 0.4em\relax Toronto, Canada: Association for Computational Linguistics, Jul. 2023, pp. 5823--5840.

\bibitem{he2022rethinking}
H.~He, H.~Zhang, and D.~Roth, ``Rethinking with retrieval: Faithful large language model inference,'' \emph{arXiv preprint arXiv:2301.00303}, 2022.

\bibitem{zhang20243a}
Y.~Zhang, S.~Gao, Y.~Huang, Z.~Yu, and K.~Tan, ``3a-cot: an attend-arrange-abstract chain-of-thought for multi-document summarization,'' \emph{International Journal of Machine Learning and Cybernetics}, pp. 1--19, 2024.

\bibitem{gupta2025autosumm}
A.~Gupta, D.~Singh, G.~A. Cowan, N.~Kadhiresan, S.~Srivastava, Y.~Sriraja, and Y.~K. Mantri, ``Autosumm: A comprehensive framework for llm-based conversation summarization,'' in \emph{Proceedings of the 63rd Annual Meeting of the Association for Computational Linguistics (Volume 6: Industry Track)}, 2025, pp. 500--509.

\bibitem{kwon2023efficient}
W.~Kwon, Z.~Li, S.~Zhuang, Y.~Sheng, L.~Zheng, C.~H. Yu, J.~Gonzalez, H.~Zhang, and I.~Stoica, ``Efficient memory management for large language model serving with pagedattention,'' in \emph{Proceedings of the 29th Symposium on Operating Systems Principles}, ser. SOSP '23.\hskip 1em plus 0.5em minus 0.4em\relax New York, NY, USA: Association for Computing Machinery, 2023, p. 611–626.

\bibitem{li2023towards}
Z.~Li, X.~Zhang, Y.~Zhang, D.~Long, P.~Xie, and M.~Zhang, ``Towards general text embeddings with multi-stage contrastive learning,'' \emph{arXiv preprint arXiv:2308.03281}, 2023.

\bibitem{martschat2017improving}
S.~Martschat and K.~Markert, ``Improving {ROUGE} for timeline summarization,'' in \emph{Proceedings of the 15th Conference of the {E}uropean Chapter of the Association for Computational Linguistics: Volume 2, Short Papers}.\hskip 1em plus 0.5em minus 0.4em\relax Valencia, Spain: Association for Computational Linguistics, Apr. 2017, pp. 285--290.

\end{thebibliography}





\appendix

\section{Discussions}

\subsection{Limitations}
Although \texttt{NTS-CoT} has shown effectiveness in mitigating hallucinations in LLM-based TLS tasks, several limitations remain to be addressed. Our approach leverages CoT prompting to guide the LLM through step-by-step reasoning, which helps reduce hallucinations in the output. However, this multi-step reasoning process increases the inference cost and overall time consumption. Moreover, the accuracy of dates in the generated timeline heavily depends on the LLM's judgment of event dates during the news summarization stage. Errors or improperly formatted outputs at this stage can negatively impact subsequent processes. Therefore, further exploration is required to ensure the quality of news summarization.

\subsection{Broader Impacts}

We proposes \texttt{NTS-CoT}, a framework for mitigating hallucinations in LLM-based news timeline summarization by leveraging element-aware and causal chain-of-thought reasoning. This work has several societal impacts. 

First, by improving the faithfulness and completeness of LLM-generated news timeline summaries, \texttt{NTS-CoT} can enhance users’ ability to understand complex event developments over time. This is particularly valuable in scenarios such as disaster response, public health emergencies, and policy evolution, where accurate and structured information is critical for timely and informed decision-making.

Second, the framework contributes to the development of more trustworthy AI systems and improves the interpretability of the summarization process. By explicitly grounding summaries in structured news elements (e.g., event, entity, loacation, date, and result) and modeling causal relationships across documents, \texttt{NTS-CoT} reduces hallucinations and improves alignment with source content. This can increase user confidence in automatically generated summaries and support more reliable information consumption.

Finally, as a prompt-based approach that does not require additional model training, \texttt{NTS-CoT} is relatively resource-efficient and flexible. This lowers the barrier to adopting more reliable summarization techniques, enabling broader use across domains and institutions with limited computational resources.

\section{Prompts}
\label{sec:appendix_prompts}
Figure~\ref{fig:prompts4Element-CoT} shows the prompts used in \emph{Element-CoT}, and Figure~\ref{fig:prompts4Causal-CoT} shows the prompts used in \emph{Causal-CoT}.

\begin{figure}
  \includegraphics[width=1\columnwidth]{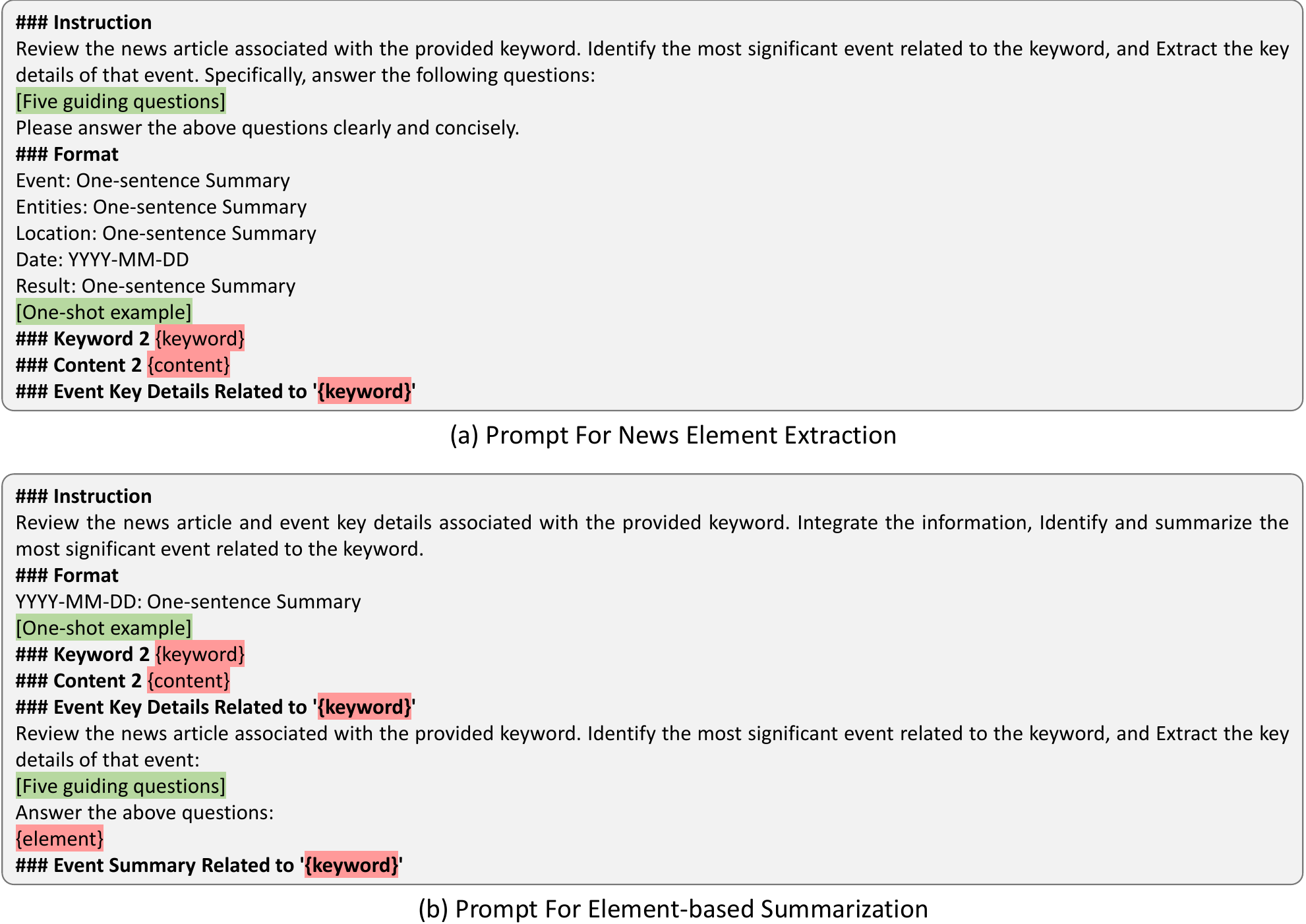}
  \caption{Prompts used in Element-CoT.}
  \label{fig:prompts4Element-CoT}
\end{figure}

\begin{figure}
  \includegraphics[width=1\columnwidth]{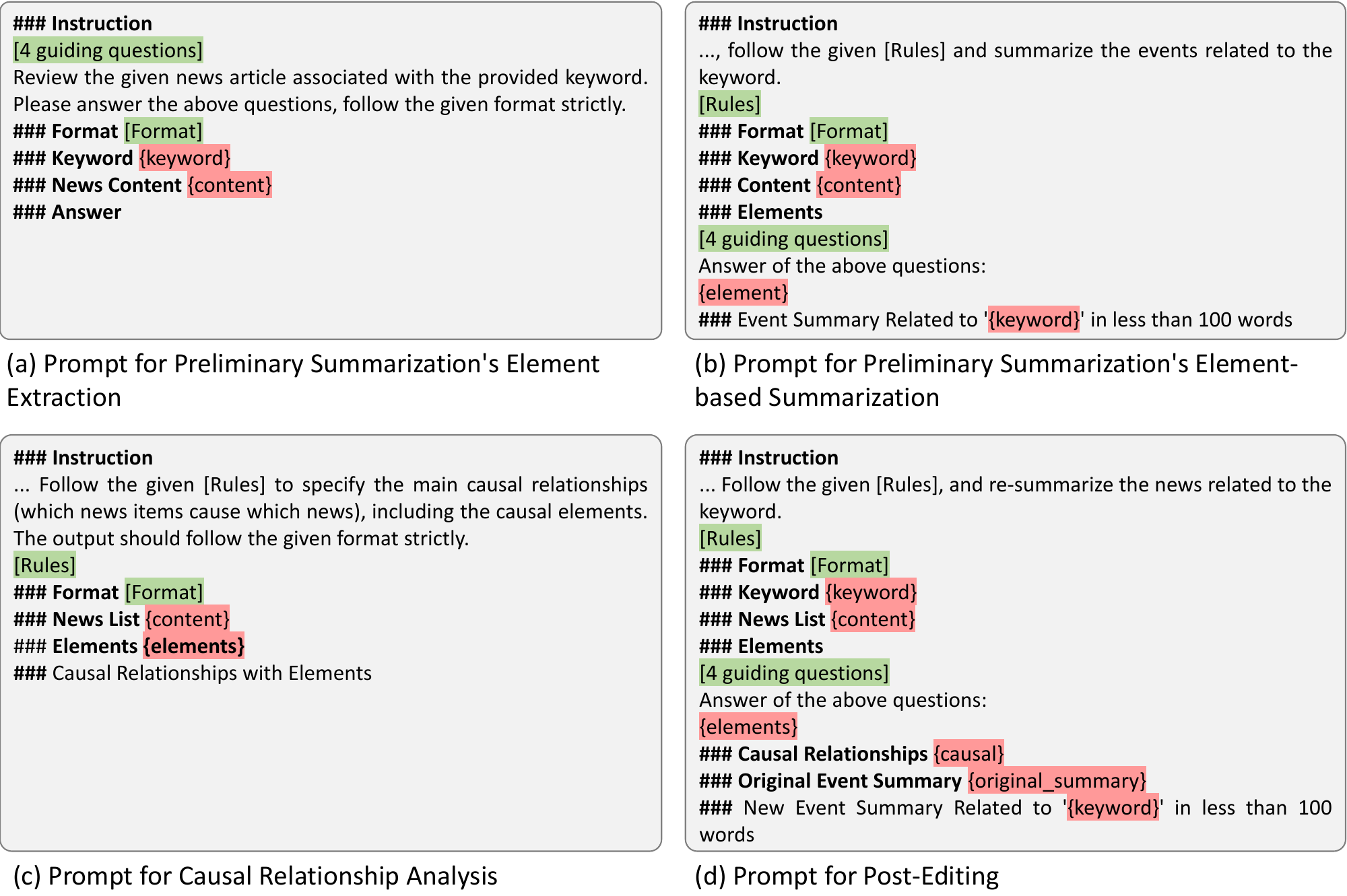}
  \caption{Prompts used in Causal-CoT.}
  \label{fig:prompts4Causal-CoT}
\end{figure}

\section{Date Selection Algorithm}
\label{sub:appendix_date_selection}
Date selection includes constructing associated date graph $G=(V,E)$, clustering event clusters $C$, and ranking the dates based on temporal saliency though $G$ and event prominence through $C$. Algorithm~\ref{alg_date} outlines the date selection process.

\begin{algorithm}
    \caption{Date selection}
    \begin{algorithmic}[1]
        \State \textbf{Input:} A set of news articles $A$, a set of news summaries $\hat{S}$, the news summaries database $D$, a set of topic query keyphrases $Q$, parameter $\alpha$, the number of timeline dates $l$.
        \State \textbf{Output:} A timeline $T$ with $l$ dates.

        \State \textbf{Associated date graph $G=(V, E)$ construction:}
        \For{$a_i$ in $A$}
            \State $t_d, t_p, t_s \leftarrow GETDATES(a_i,Q)$
            \State $V \leftarrow V \cup \{v_{t_d}, v_{t_p}, v_{t_s}\}$
            \State $E \leftarrow E \cup \{[v_{t_p},v_{t_p}], [v_{t_p},v_{t_d}], [v_{t_p},v_{t_s}]\}$
        \EndFor    
        \State \textbf{Event Clustering:}
        \For{$\hat{s}_{a_i}$ in $\hat{S}$}
            \State $Events,D \leftarrow RETRIEVE(\hat{s}_{a_i},D)$
            \For{$\hat{s}_{a_j}$ in $Events$}
            \If{$ISSAMEEVENT(\hat{s}_{a_i},\hat{s}_{a_j})$}
                \State $C \leftarrow CLUSTERING(\hat{s}_{a_i},\hat{s}_{a_j})$
            \EndIf
        \EndFor
        \EndFor
        \State \textbf{Date ranking:}
            \For{$t_i$ in $G,C$}
                \State $score(t_i) \leftarrow \alpha \cdot d_{in}\left(v_{t_i} \right)+\left( 1-\alpha  \right)\cdot \left| C\left(t_i \right) \right|$
            \EndFor
            \State $T \leftarrow RANK(score(t_i),l)$
            \State $T \leftarrow SORTBYTIME(T)$
        \State \textbf{Return} $T$
    \end{algorithmic}
    \label{alg_date}
\end{algorithm}

\section{Experiment Details}

\subsection{Implementation Details}
We employ Llama3-8B-Instruct \cite{dubey2024llama} as the foundational language model in the experiment. We leverage the vLLM library \cite{kwon2023efficient} for efficient LLM inference. All experiments are performed on an A100 80GB GPU. For the similarity retrieval step in \emph{Event Clustering}, we use the gte-large model \cite{li2023towards} for text embedding. In the \emph{Date Ranking} step, the parameter $\alpha$ is set to 0.5. During the chunking step, the value of ${{M}_{B}}$ is set to 20.

\subsection{Evaluation Metrics}
We evaluate the results using AR-1, AR-2, and Date F1.
\begin{itemize}
    \item \textbf{Alignment-based ROUGE F1-score (AR1 \& AR2).} This metric \cite{martschat2017improving} aligns the generated summaries with the ground-truth summaries based on their similarity and the temporal distance between their dates, and then computes the textual overlap using ROUGE-1 and ROUGE-1. The higher AR score indicates greater text overlap, reflecting fewer hallucinations of unfaithful content and information omission.
    \item \textbf{Date F1-score (Date F1).} This metric computes the F1 score for dates in the generated timeline compared to the ground-truth timeline, indirectly reflecting the faithfulness of the date content.
\end{itemize}

\subsection{Baselines Details}
We compare the performance of \emph{NTS-CoT} against several baselines, including both extractive \cite{martschat-markert-2018-temporally, gholipour-ghalandari-ifrim-2020-examining, la2021summarize, li-etal-2021-timeline} and abstractive \cite{steen-markert-2019-abstractive} non-LLM-based methods, as well as state-of-the-art LLM-based approaches \cite{hu-etal-2024-moments}. Due to the limited availability of open-source LLM-based methods, in addition to adopting an existing advanced LLM-TLS approach, we also employ different prompting strategies for LLM-based summarization.

\begin{itemize}
\item \textbf{MARTSCHAT}
 \cite{martschat-markert-2018-temporally} adapts the submodular function model to select a combination of sentences that maximizes content coverage, textual and temporal diversity.
\item \textbf{DATEWISE}
 \cite{gholipour-ghalandari-ifrim-2020-examining} employs a supervised method to select dates and then applies a heuristic to choose sentences corresponding to the dates.
\item \textbf{CLUST}
 \cite{gholipour-ghalandari-ifrim-2020-examining} uses Markov Clustering to group articles and then applies CENTROID-OPT to summarize each cluster.
\item \textbf{SDF}
 \cite{la2021summarize} summarizes dates by first using an unsupervised summarization algorithm to extract per-date summary sentences, followed by date selection.
\item \textbf{EGC}
 \cite{li-etal-2021-timeline} models news articles as an event graph, using time-aware optimal transport to compress the graph into a salient sub-graph from which key events are extracted.
\item \textbf{STEEN}
 \cite{steen-markert-2019-abstractive} utilizes an unsupervised abstractive TLS system to generate readable sentences as output.
\item \textbf{LLM-TLS}
 \cite{hu-etal-2024-moments} leverages LLMs for news summaries generation and event clustering, then selects sentences from the clusters as date summaries. 
\item \textbf{Zero-shot.} This method retains the date selection design while prompting the LLM to generate summaries using a simple zero-shot instruction.
\item \textbf{One-shot.} A single example is provided, but no multi-step reasoning is performed.
\item \textbf{Standard-CoT.} No example is provided, but multi-step reasoning is conducted.
\end{itemize}

\subsection{Human Evaluation Criteria and Case Study}
\label{sec:appendix_human}
To further evaluate our approach, we randomly select 30 timelines generated by LLM-TLS and \texttt{NTS-CoT} across three datasets for human evaluation. To assess hallucinations in timeline summaries, human evaluators compare outputs for each date-event summary occurring on the same date within a timeline and select the one with fewer hallucinations based on two criteria: Faithfulness and Completeness. For pairwise evaluation, date-event summaries must satisfy the following conditions: (1) belong to the same timeline task, (2) occur on the same date, and (3) have a date that appears in the ground truth timeline. 
We randomly recruit three students as volunteers for the human evaluation. We explain the human evaluation criteria and the purpose of the experimental data for them. The timeline results are anonymized, and the final preference scores were based on the average of the volunteers' assessments.

The details for the two human evaluation criteria are shown below:
\begin{itemize}
\item \textit{Faithfulness}: Which date-event summary occurring on the same date in the timelines is more consistent with the information in the news article? Focus on the summary that correctly and accurately conveys the meaning of the news article.
\item \textit{Completeness}: Which date-event summary occurring on the same date in the timelines includes all relevant elements and has the fewest missing important details? Focus on the summary that contains fewer unimportant or irrelevant elements.
\end{itemize}

As shown in Table~\ref{tab:case}, we provide a case study for the human evaluation analysis comparing the output date-event summaries of LLM-TLS (Llama3-8B-Instruct) and \texttt{NTS-CoT} (Llama3-8B-Instruct) given a specific text description. \texttt{NTS-CoT} identifies the key elements and correctly output ``\textit{Bill Clinton is acquitted in the Senate impeachment trial}'', while the LLM-TLS output contains a hallucination, mistakenly stating ``\textit{The US Senate voted to convict President Bill Clinton}.''

\begin{table}[h]
  \centering
  \caption{A case study for the human evaluation analysis presenting the ground-truth date-event summaries alongside the outputs of LLM-TLS and NTS-CoT.}
    \begin{tabularx}{\columnwidth}{l|X}
    \hline
    \makecell[tl]{Ground-\\truth} & \{"1999-02-12", ["Senate trial ends with an \textcolor{red}{acquittal}.", "The vote on the perjury charge is 55 to 45 and the obstruction of justice charge is split 50-50.", "A two-thirds majority, or 67 votes, was required for conviction."]\} \\
    \hline
    \makecell[tl]{LLM-TLS} & \{"1999-02-12": ["The US Senate voted to \textcolor{red}{convict} President Bill Clinton on charges of perjury and obstruction of justice, with nine Republican senators breaking ranks to vote with the Democrats."]\} \\
    \hline
    \makecell[tl]{NTS-CoT} & \{"1999-02-12": ["Bill Clinton is \textcolor{red}{acquitted} of perjury and obstruction of justice charges in the Senate impeachment trial, bringing an end to the trial and sparking a media frenzy surrounding the scandal."]\} \\
    \hline
    \end{tabularx}%
  \label{tab:case}%
\end{table}%

\subsection{Robustness Across Different LLMs}

To evaluate the generalization and robustness of \texttt{NTS-CoT} across different LLM backbones, we conduct experiments using various models. The results are reported in the table. Overall, \texttt{NTS-CoT} achieves the best performance when utilizing Llama3-8B-Instruct as the base model, which is adopted as the default backbone in this work. Moreover, comparison with the Zero-shot setting further demonstrates that our method achieves consistent performance improvements on the TLS task.

\begin{table}[H]
  \centering
  \caption{Experimental results of different LLMs on T17.}
  \resizebox{\textwidth}{!}{
    \begin{tabular}{clccccccccc}
    \hline
    \multirow{2}[2]{*}{\textbf{Model}} & \multicolumn{1}{c}{\multirow{2}[2]{*}{\textbf{Method}}} & \multicolumn{3}{c}{\textbf{T17}} & \multicolumn{3}{c}{\textbf{Crisis}} & \multicolumn{3}{c}{\textbf{Entities}} \bigstrut[t]\\
          &       & \textbf{AR-1} & \textbf{AR-2} & \textbf{Date F1} & \textbf{AR-1} & \textbf{AR-2} & \textbf{Date F1} & \textbf{AR-1} & \textbf{AR-2} & \textbf{Date F1} \bigstrut[b]\\
    \hline
    \multirow{2}[2]{*}{\textbf{ChatGLM3-6B}} & Zero-shot & 0.092  & 0.023  & 0.501  & 0.073  & 0.018  & 0.289  & 0.051  & 0.016  & 0.192  \bigstrut[t]\\
          & NTS-CoT & 0.093  & 0.023  & 0.524  & 0.080  & 0.020  & 0.301  & 0.055  & 0.016  & 0.217  \bigstrut[b]\\
    \hline
    \multirow{2}[2]{*}{\textbf{Qwen2.5-7B}} & Zero-shot & 0.106  & 0.025  & 0.510  & 0.087  & 0.021  & 0.273  & 0.062  & 0.021  & 0.198  \bigstrut[t]\\
          & NTS-CoT & 0.109  & 0.028  & 0.526  & 0.088  & 0.022  & 0.303  & 0.072  & 0.025  & \textbf{0.245 } \bigstrut[b]\\
    \hline
    \multirow{2}[2]{*}{\textbf{Llama3-8B-Instruct}} & Zero-shot & 0.097  & 0.027  & 0.504  & 0.068  & 0.016  & 0.271  & 0.048  & 0.016  & 0.193  \bigstrut[t]\\
          & NTS-CoT & \textbf{0.116}  & \textbf{0.032}  & \textbf{0.541}  & \textbf{0.094 } & \textbf{0.024}  & \textbf{0.305 } & \textbf{0.075 } & \textbf{0.027 } & 0.230  \bigstrut[b]\\
    \hline
    \end{tabular}%
  \label{tab:robustness}%
  }
\end{table}%



\end{document}